\documentclass[11pt,a4paper]{article}
\usepackage[hyperref]{eacl2021}
\usepackage{times}
\usepackage{latexsym}

\usepackage{microtype}

\usepackage{times}
\usepackage{url}
\usepackage{latexsym}
\usepackage{arabtex}
\usepackage{amsmath,amssymb,amsfonts}
\usepackage{graphicx}
\usepackage{textcomp}
\usepackage{subcaption}
\usepackage{utf8}
\usepackage{multirow}
\usepackage{enumitem}

\usepackage{colortbl}
\usepackage{tikz}

\makeatletter
\def\thickhline{%
  \noalign{\ifnum0=`}\fi\hrule \@height \thickarrayrulewidth \futurelet
   \reserved@a\@xthickhline}
\def\@xthickhline{\ifx\reserved@a\thickhline
               \vskip\doublerulesep
               \vskip-\thickarrayrulewidth
             \fi
      \ifnum0=`{\fi}}
\makeatother

\newlength{\thickarrayrulewidth}
\aclfinalcopy 

\renewenvironment{abstract}%
		 {\centerline{\large\bf Abstract}%
		  \begin{list}{}%
		     {\setlength{\rightmargin}{0.6cm}%
		      \setlength{\leftmargin}{0.6cm}}%
		   \item[]\ignorespaces
		   \smalll
		   }%
		 {\unskip\end{list}}
		 
\title{Empathetic BERT2BERT Conversational Model:\\ Learning Arabic Language Generation with Little Data}

\author{Tarek Naous, Wissam Antoun, Reem A. Mahmoud, and Hazem Hajj \\
  Department of Electrical and Computer Engineering \\
  American University of Beirut \\
  Beirut, Lebanon \\
  {\tt \{tnn11,wfa07,ram79,hh63\}@aub.edu.lb}
  }

\date{}

\begin{document}
\maketitle

\begin{abstract}
Enabling empathetic behavior in Arabic dialogue agents is an important aspect of building human-like conversational models. While Arabic Natural Language Processing has seen significant advances in Natural Language Understanding (NLU) with language models such as AraBERT, Natural Language Generation (NLG) remains a challenge. The shortcomings of NLG encoder-decoder models are primarily due to the lack of Arabic datasets suitable to train NLG models such as conversational agents. To overcome this issue, we propose a transformer-based encoder-decoder initialized with AraBERT parameters. By initializing the weights of the encoder and decoder with AraBERT pre-trained weights, our model was able to leverage knowledge transfer and boost performance in response generation. To enable empathy in our conversational model, we train it using the ArabicEmpatheticDialogues dataset and achieve high performance in empathetic response generation. Specifically, our model achieved a low perplexity value of 17.0 and an increase in 5 BLEU points compared to the previous state-of-the-art model. Also, our proposed model was rated highly by 85 human evaluators, validating its high capability in exhibiting empathy while generating relevant and fluent responses in open-domain settings.
\end{abstract}


\section{Introduction}

\begin{figure}[t]
    \centering
    \includegraphics[width=\linewidth]{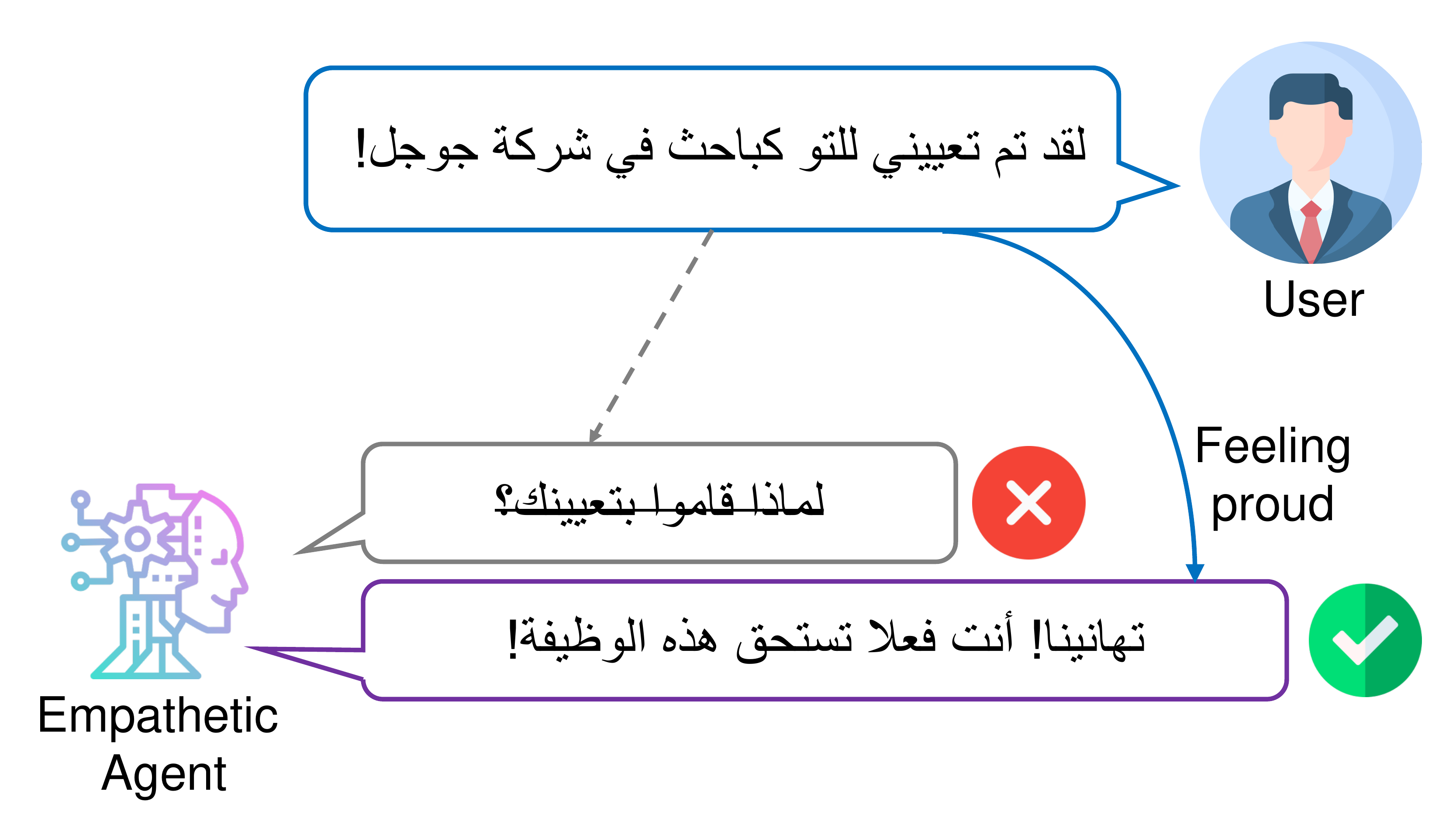}
    \caption{Example of empathetic behavior in a conversational agent.}
    \label{fig:example-intro}
\end{figure}

Conversational models with empathetic responding capabilities are crucial in making human-machine interactions closer to human-human interactions, as they can lead to increased engagement, more trust, and reduced frustration \cite{E15}. These characteristics are highly desirable in open-domain conversational models as they can boost user satisfaction and make chatbots look less boorish. While empathy can be attributed to a range of behaviors, it can be generally described as the innate human capacity of relating to another person’s feelings and making sense of their emotional state \cite{E8}. An important factor towards developing human-like dialogue agents is enabling their empathetic capability \cite{I1}. To this end, there has been a significant interest in developing empathetic conversational models \cite{I2,I3,I4,mpath}. These models infer the emotions of a human user and provide a suitable empathetic response. The desired behavior of an empathetic conversational agent is illustrated in Figure~\ref{fig:example-intro}, where the empathetic agent recognizes that the user is feeling proud and, thus, generates an empathetic response that congratulates the user with enthusiasm.

Recent work in open-domain empathetic conversational models have adopted neural-based sequence generation approaches \cite{E2}. These approaches are based on encoder-decoder neural network architectures such as Sequence-to-Sequence (Seq2Seq) recurrent neural network models \cite{E17} or transformers \cite{E3}. Despite the significant work done in this direction, the focus so far has been mostly on the English language with fewer efforts being directed towards low-resource languages, such as Arabic. The first dataset for Arabic utterances and empathetic responses was recently introduced by \newcite{naous}, where a Bidirectional Long Short-Term Memory (Bi-LSTM) Seq2Seq model was trained on the dataset. However, the model proposed by \newcite{naous} delivered sub-optimal performance due to the limited size of the dataset. The additional challenges in developing neural-based empathetic conversational models for Arabic is the lack of open-domain conversational data that can be used for pre-training \cite{dailydialog}, and thus no availability of pre-trained conversational models that can be used directly for fine-tuning \cite{dialogpt}.

To address the challenges of small dataset size and lack of conversational resources, in terms of datasets and pre-trained models, we propose a transformer-based encoder-decoder model initialized with AraBERT \cite{arabert} pre-trained weights. Our work extends the English BERT2BERT architecture \cite{bert2bert} to Arabic response generation. We fine-tune our proposed model on the limited-sized dataset of empathetic responses in Arabic \cite{naous}. By using the pre-trained weights of the AraBERT language model to initialize the encoder and decoder, our proposed BERT2BERT model is expected to leverage knowledge transfer and show enhanced performance in empathetic response generation compared to the baseline Bi-LSTM model proposed by \newcite{naous}.

The rest of this paper is organized as follows: Section~\ref{sec:related-work} reviews the recent literature on empathetic conversational models in both English and Arabic. Our proposed BERT2BERT approach for empathetic response generation is presented in Section~\ref{sec:prop-method}, including the dataset and pre-processing steps. Section~\ref{sec:experiment-results} analyzes the performance of our model and compares its results to several benchmark models. Concluding remarks and future directions are presented in Section~\ref{sec:conc}.

\section{Related Work}
\label{sec:related-work}

\subsection{English Empathetic Conversational Models}

The interest in enabling empathy in conversational agents has increased over the last few years with the introduction of the EmpatheticDialogues dataset by \newcite{E2}. EmpatheticDialogues is a crowdsourced dataset of open-domain conversations where a group of workers was instructed to select an emotion, describe a situation where they have felt that way, and carry out a conversation related to the emotion. The authors used the conversations collected to train retrieval-based and generative-based models, which showed higher levels of empathy in their responses compared with models trained on spontaneous conversational data gathered from the Internet. 

The release of EmpatheticDialogues \cite{E2} stimulated further research in this area with multiple attempts in the literature to improve the empathetic capability of conversational models. \newcite{E17} formulated the empathetic responding task as a reinforcement learning problem. The approach named “Sentiment look-ahead” employs a Seq2Seq policy model with Gated Recurrent Units to generate an empathetic response based on an input utterance and updates the policy using the REINFORCE method. \newcite{E3} fined-tuned a GPT model on the EmpatheticDialogues dataset. The GPT model was pre-trained on the BooksCorpus \cite{bookscorpus} dataset, improving the NLU capability of the model, as well as on the PersonaChat \cite{personachat} dataset, allowing the model to have improved performance on response generation.

\begin{figure*}[t]
    \centering
    \includegraphics[scale=0.4]{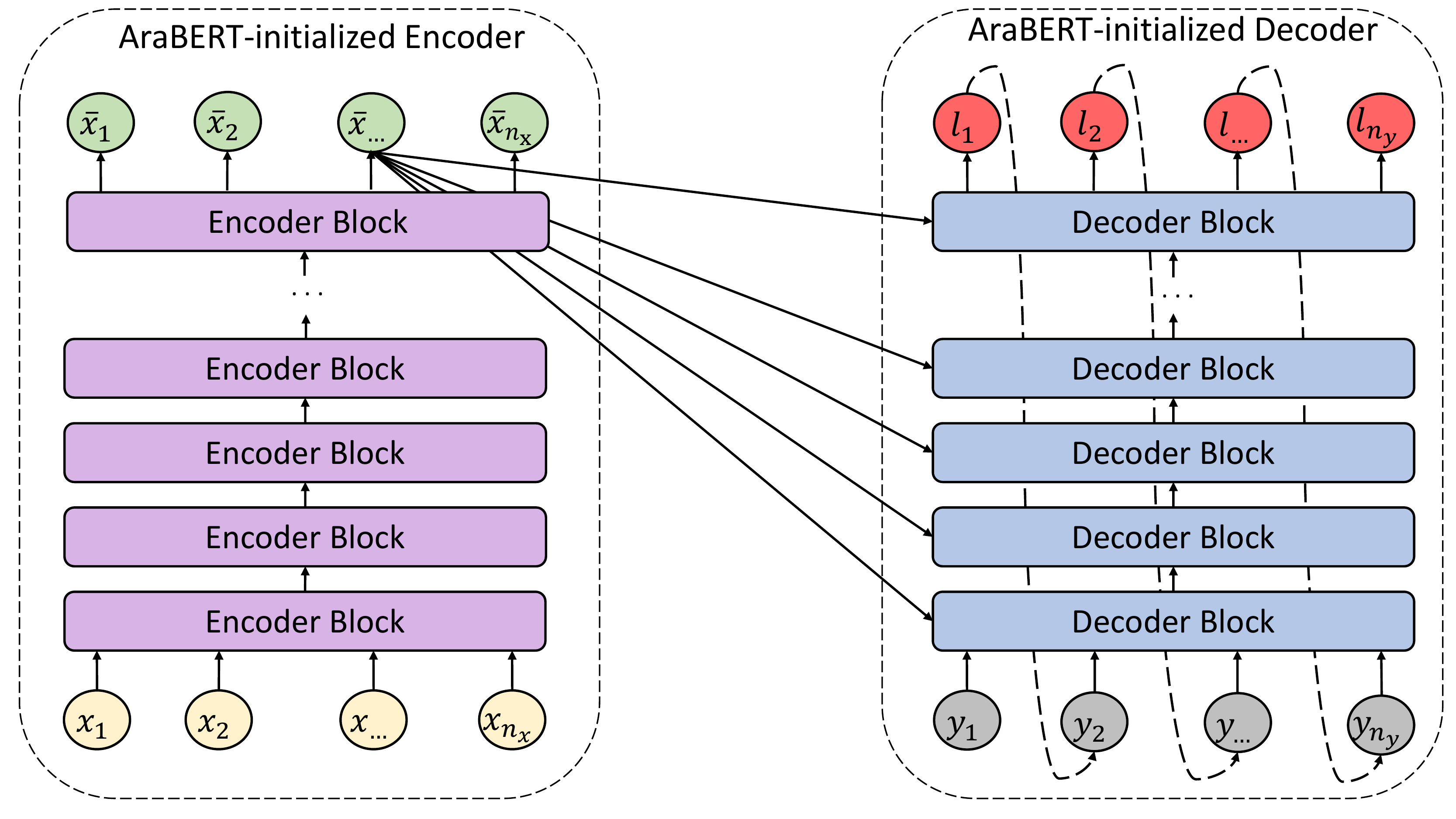}
    \caption{Architecture of the proposed BERT2BERT model initialized with AraBERT checkpoints for Arabic empathetic response generation.}
    \label{fig:bert2bertarch}
\end{figure*}

\subsection{Arabic Empathetic Conversational Models}
While many works have focused on enabling empathetic capabilities in conversational models for English, there are much fewer attempts to build similar models for Arabic. In general, research on Arabic conversational models is still in its infancy mainly due to the complexity of the language, and the lack of resources and pre-trained models that are available in abundance for English. Despite the availability of Arabic pre-trained language models such as hULMonA \cite{hulmona} and AraBERT \cite{arabert}, which have proven useful for Arabic NLU tasks, the lack of pre-trained models for Arabic NLG makes the development of neural-based Arabic conversational models a challenging task.  Hence, existing works on Arabic chatbots have mainly focused on retrieval-based methods \cite{botta} or rule-based approaches \cite{arabchat,ollobot}. While these approaches work well on task-oriented objectives, they are limited by the size of manually crafted rules they follow or the richness of the database they can retrieve responses from. This makes it difficult for such types of models to operate well in open-domain conversational settings, where generative neural-based models would be more suitable.

Recently, the first empathy-driven Arabic conversational model was proposed by \newcite{naous} that released ArabicEmpatheticDialogues, a dataset of Arabic utterances and their corresponding empathetic responses. The authors trained a Seq2Seq model with bidirectional LSTM units on the dataset. While the model succeeded in generating empathetic responses, it showed an average Relevance score which indicates that the responses can sometimes go off-topic and may not be suitable responses for the emotional context of the input utterance. The limitations of this work were mainly due to the limited size of the dataset.

In this work, we adopt the BERT2BERT architecture \cite{bert2bert} and leverage the pre-trained AraBERT \cite{arabert} model to improve the performance of empathetic Arabic conversational models.

\section{Proposed Method}
\label{sec:prop-method}

\subsection{Proposed BERT2BERT Model}
Our proposed model for Arabic empathetic response generation is a transformer-based Seq2Seq model \cite{trans-enc-dec}, which has been shown to boost performance on a several Seq2Seq tasks \cite{seq2seq-task1,seq2seq-task2}. However, such an architecture would require massive pre-training before being fine-tuned on the desired task \cite{pretraining}. It was shown by \newcite{bert2bert} that warm-starting the transformer-based encoder-decoder model with the checkpoints of a pre-trained encoder (e.g. BERT) allows the model to deliver competitive results in sequence generation tasks while skipping the costly pre-training. Inspired by this idea, and due to the unavailability of Arabic conversational datasets that can be used for pre-training, we adopt the BERT2BERT architecture \cite{bert2bert}, and warm-start the encoder and decoder with the AraBERT checkpoint \cite{arabert}. The encoder-decoder attention is randomly initialized.  The architecture of the proposed model is illustrated in Figure~\ref{fig:bert2bertarch}.

The input to the proposed model is a sequence $x = [x_1, x_2, \dots, x_{n_x}]$ of one-hot representations with a length of $n_x$ tokens, chosen to be 150. This sequence is fed as input to an AraBERT initialized encoder. At the decoder side, the model generates an empathetic response represented by a sequence $y = [y_1, y_2, \dots, y_{n_y}]$, where the maximum output length $n_y$ is also specified to be 150. We optimize the log-likelihood loss over the output tokens.

\begin{table*}[ht]
\setcode{utf8}
\centering
\begin{tabular}{cr}
\thickhline
Emotion   & \multicolumn{1}{c}{\textbf{Excited}}                                                             \\ \cline{1-1}
Utterance & \small{\< في الأسبوع الماضي كنا نشاهد كأس العالم، وها هي كرواتيا تغلبت على البلد المضيف، روسيا>} \\ \cline{1-1}
Response  & \small{\< رأيت تلك اللعبة. لقد فازوا بركلات الترجيح!>}                                           \\
\hline \hline 
Emotion   & \multicolumn{1}{c}{\textbf{Furious}}                                                             \\ \cline{1-1}
Utterance & \small{\<تخيل، لقد طلبت البطاطس المقلية و لكن تم تقديم البرغر بدلا من ذلك!>}                     \\ \cline{1-1}
Response  & \small{\<إنها خدمة سيئة للغاية. ماذا فعلت؟ هل اشتكيت للمدير؟ >}                                  
\\ \hline \hline 
Emotion   & \multicolumn{1}{c}{\textbf{Embarrassed}}                                                         \\ \cline{1-1}
Utterance & \small{\<خرجت في نهاية الأسبوع الماضي و تعرضت لحادث كبير. خمن ماذا جرى>}                         \\ \cline{1-1}
Response  & \small{\<هل أنت بخير؟ عليك أن تخبرني بما حدث>}                                                   \\ \thickhline
\end{tabular}
\caption{Samples of utterances and empathetic responses from the ArabicEmpatheticDialogues dataset for three emotion labels: Excited, Furious, and Embarrassed}
\label{tab:dataset-samples}
\end{table*}

To generate empathetic responses from our model, we adopt the Top-K Sampling scheme \cite{topksampling} where, at each time step, the model randomly samples the K most likely candidates from the probability distribution of all words in the vocabulary. This decoding strategy has been found more effective than conventional approaches such as beam search, which tends to yield common responses found repetitively in the training set or similar, slightly-varying versions of the same high-likelihood sequences \cite{decodingstrategies}.

\subsection{Dataset}

\begin{table}[h]
\centering
\small
\begin{tabular}{c|c}
\textbf{Grouped Emotion Labels}           & \textbf{Complete Emotion Labels}                \\ \thickhline
                                    & \cellcolor[HTML]{EFEFEF}Excited      \\ \cline{2-2} 
                                    & Proud                                \\ \cline{2-2} 
                                    & \cellcolor[HTML]{EFEFEF}Grateful     \\ \cline{2-2} 
                                    & Hopeful                              \\ \cline{2-2} 
                                    & \cellcolor[HTML]{EFEFEF}Confident    \\ \cline{2-2} 
                                    & Joyful                               \\ \cline{2-2} 
                                    & \cellcolor[HTML]{EFEFEF}Content      \\ \cline{2-2} 
                                    & Prepared                             \\ \cline{2-2} 
\multirow{-9}{*}{\textbf{Joy}}      & \cellcolor[HTML]{EFEFEF}Anticipating \\ \thickhline
                                    & Caring                               \\ \cline{2-2} 
                                    & \cellcolor[HTML]{EFEFEF}Sentimental  \\ \cline{2-2} 
                                    & Trusting                             \\ \cline{2-2} 
                                    & \cellcolor[HTML]{EFEFEF}Faithful     \\ \cline{2-2} 
\multirow{-5}{*}{\textbf{Love}}     & Nostalgic                            \\ \thickhline
                                    & \cellcolor[HTML]{EFEFEF}Surprised    \\ \cline{2-2} 
\multirow{-2}{*}{\textbf{Surprise}} & Impressed                            \\ \thickhline
                                    & \cellcolor[HTML]{EFEFEF}Sad          \\ \cline{2-2} 
                                    & Lonely                               \\ \cline{2-2} 
                                    & \cellcolor[HTML]{EFEFEF}Guilty       \\ \cline{2-2} 
                                    & Disappointed                         \\ \cline{2-2} 
                                    & \cellcolor[HTML]{EFEFEF}Devastated   \\ \cline{2-2} 
                                    & Embarrassed                          \\ \cline{2-2} 
\multirow{-7}{*}{\textbf{Sadness}}  & \cellcolor[HTML]{EFEFEF}Ashamed      \\ \thickhline
                                    & Angry                                \\ \cline{2-2} 
                                    & \cellcolor[HTML]{EFEFEF}Annoyed      \\ \cline{2-2} 
                                    & Furious                              \\ \cline{2-2} 
                                    & \cellcolor[HTML]{EFEFEF}Disgusted    \\ \cline{2-2} 
\multirow{-5}{*}{\textbf{Anger}}    & Jealous                              \\ \thickhline
                                    & \cellcolor[HTML]{EFEFEF}Afraid       \\ \cline{2-2} 
                                    & Terrified                            \\ \cline{2-2} 
                                    & \cellcolor[HTML]{EFEFEF}Anxious      \\ \cline{2-2} 
\multirow{-4}{*}{\textbf{Fear}}     & Apprehensive                          \\ \thickhline   
\end{tabular}
\caption{Grouping of emotion labels in the  ArabicEmpatheticDialogues dataset as per Parrott's characterization of tree-structured emotions \cite{emotions}.}
\label{tab:emotion-grouping}
\end{table}

We use the ArabicEmpatheticDialogues dataset \cite{naous} which was translated from the English version introduced by \newcite{E2}. ArabicEmpathicDialogues contains 36,628 samples of speaker utterances and their corresponding empathetic responses in Arabic. Each sample is also labeled with the emotion of the speaker's utterance. Three examples from the dataset for three different emotion labels are provided in Table~\ref{tab:dataset-samples}. By training a sequence generation model on the samples of utterances and their corresponding responses from the dataset, the model will be able to infer the emotions in input utterances and provide suitable empathetic responses. Thus, the empathetic capability of the model would be enhanced.

The dataset is originally labeled with 32 emotion labels, many of which are very similar such as “joyful” and “content”, or “angry” and “furious”. To reduce the number of classes, we follow the tree-structured list of emotions defined by \newcite{emotions} to map the 32 emotion labels to their 6 primary emotions which are “Joy”, “Surprise”, “Love”, “Surprise”, “Anger”, and “Fear”. This grouping is shown in Table~\ref{tab:emotion-grouping}.

\begin{table}[t]
\setcode{utf8}
\centering
\begin{tabular}{r}
\hline
\multicolumn{1}{l}{\textbf{Pre-Segmentation}}                 \\
\small{\<أنا فخور جدا بإبنتي لقد تخرجت للتو من كلية الهندسة>} \\ \hline
\multicolumn{1}{l}{\textbf{Post-Segmentation}}                \\
\small{\<أنا فخور جد+ا  ب+ إبن +ت +ي  لقد تخرج +ت  >}         \\
\small{\<ل+ ال+ تو من كلي +ة ال+ هندس +ة>}                    \\ \hline
\end{tabular}
\caption{Example of an Arabic utterance segmentation using Farasa.}
\label{tab:farasa}
\end{table}

To reduce lexical sparsity, the utterances and responses in the dataset are segmented using the Farasa segmenter \cite{farasa}. Given the morphological complexity of the Arabic language, segmentation is an important pre-processing step that can greatly enhance the performance of neural-based sequence generation models. An example of this process is shown in Table~\ref{tab:farasa}. By performing segmentation, the vocabulary size is drastically reduced from 47K tokens to around 13K tokens.

\begin{figure}[ht]
    \centering
    \includegraphics[width=\linewidth]{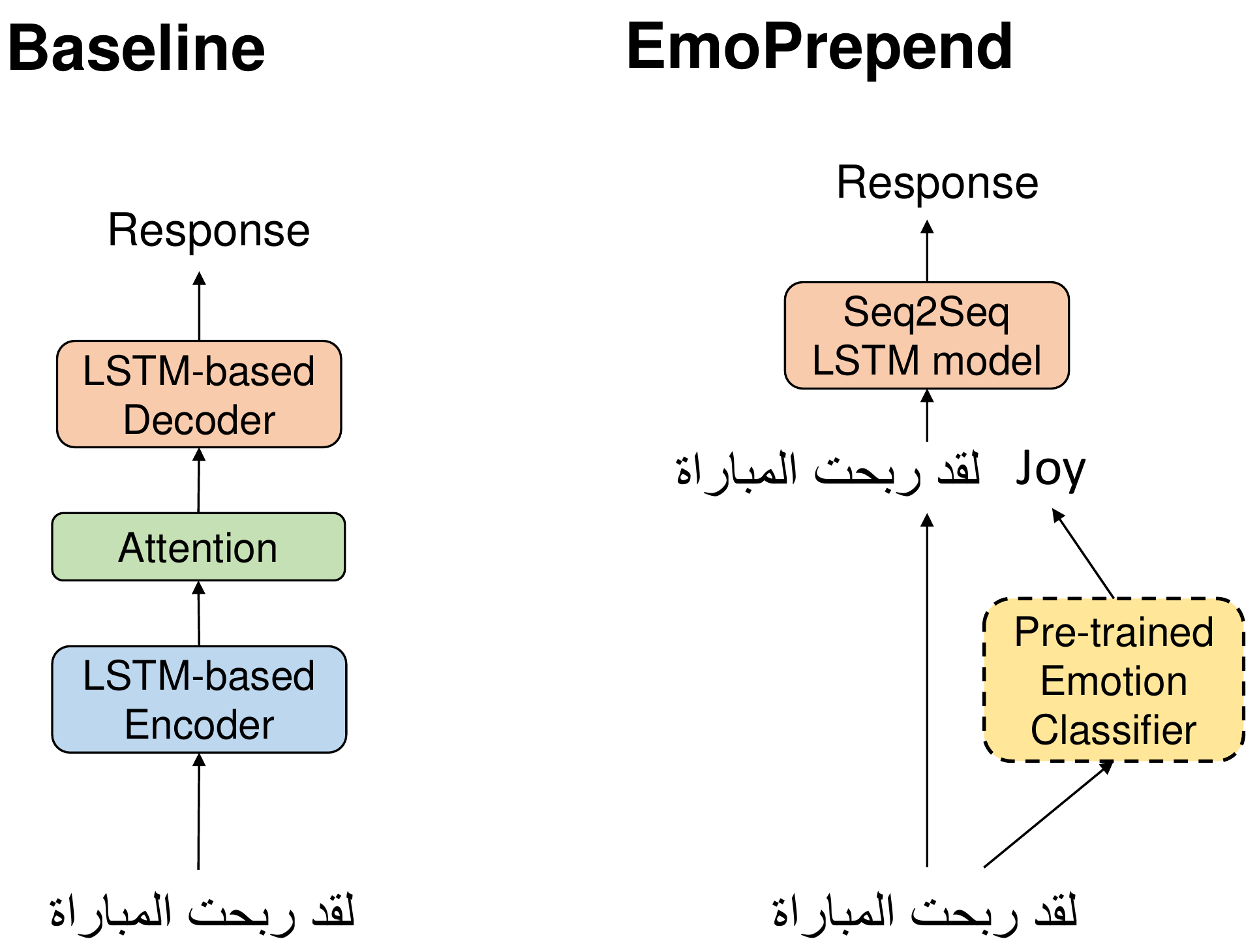} 
    \caption{Architectures of the Baseline and EmoPrepend models used for comparative evaluation against the proposed BERT2BERT model.}
    \label{fig:benchmark-models}
\end{figure}

\section{Experiments \& Results} 
\label{sec:experiment-results}

We evaluate the proposed BERT2BERT model in comparison to three benchmark models. We conduct numerical as well as human evaluation of the different conversational models.

\subsection{Benchmark Models}
We train several neural-based sequence generation models on the ArabicEmpatheticDialogues dataset and consider them as benchmarks for performance comparison. The benchmark models are denoted as follows:

\textbf{Baseline:} The baseline model, illustrated in Figure~\ref{fig:benchmark-models}, is a Seq2Seq Bi-LSTM model with Attention following the prior state-of-the-art model proposed by \newcite{naous}.

\textbf{EmoPrepend:} In this setup, illustrated in Figure~\ref{fig:benchmark-models}, we prepend the emotion label to each utterance before feeding it as input to the baseline model described above, and we denote this approach as EmoPrepend. This allows us to add supervised information to the data, without having to introduce any modifications to the architecture. The existing emotion labels have been prepended to the utterances in the train and validation sets. For the test set and at inference, we fine-tune AraBERT for emotion classification using the utterances and their labels in the dataset. The fine-tuned AraBERT model is then used as an external predictor to classify the emotion in the utterance and prepend it as a token before being used as an input to the EmoPrepend model. We note that the step of grouping emotion labels into 6 main labels, as discussed in Section 3, makes the emotion classification task easier.

\textbf{BERT2BERT-UN:} which stands for BERT2BERT-Uninitialized. This model is a regular transformer-based encoder-decoder model that shares the same architecture of the BERT2BERT model shown in Figure~\ref{fig:bert2bertarch}, but \textbf{is not} initialized with AraBERT pre-trained weights.

\subsection{Experimental Setup}
The proposed BERT2BERT model was developed using the Huggingface transformers library\footnote{https://github.com/huggingface/transformers}. We train the model for 5 epochs with a batch size of 32\footnote{https://github.com/aub-mind/Arabic-Empathetic-Chatbot}. Model training was done on a 16GB V100 NVidia GPU. The Baseline Bi-LSTM Seq2Seq \cite{naous}, EmoPrepend, and BERT2BERT-UN benchmark models were developed using the OpenNMT Library \cite{opennmt}.

\textbf{Dataset Partitioning}: All models were trained and evaluated on common data splits of the ArabicEmpatheticDialogues. We randomly partitioned the dataset into 90\% training, 5\% validation, and 5\% testing using a seed of 42.

\begin{table}[t]
\centering
\begin{tabular}{lcc}
\thickhline
\multicolumn{1}{c}{\textbf{Model}} & \textbf{PPL} & \multicolumn{1}{l}{\textbf{BLEU}} \\ \hline
Baseline \small{\cite{naous}}      & 38.6         & 0.5                               \\
EmoPrepend                         & 24.1         & 3.16                              \\
BERT2BERT-UN                        & 159.8        & 0.1                               \\
BERT2BERT                          & \textbf{17.0}         & \textbf{5.58}                              \\ \hline \hline
\end{tabular}
\caption{Performance of the models on the test set in terms of PPL and BLEU score.}
\label{tab:metrics}
\end{table}

\begin{table*}[t]
\centering
\begin{tabular}{llccccc}
\thickhline
\multicolumn{1}{c}{\textbf{Model}} & \textbf{} & \multicolumn{1}{l}{\textbf{Empathy}} & \multicolumn{1}{l}{\textbf{}} & \multicolumn{1}{l}{\textbf{Relevance}} & \multicolumn{1}{l}{\textbf{}} & \multicolumn{1}{l}{\textbf{Fluency}} \\ \thickhline
Baseline \cite{naous}              &           & 2.04                                 &                               & 1.68                                   &                               & 2.44                                 \\
EmoPrepend                         &           & 2.81                                 &                               & 2.18                                   &                               & 3.28                                 \\
BERT2BERT                          &           & \textbf{4.0}                         &                               & \textbf{3.59}                          &                               & \textbf{4.30}                        \\ \hline \hline
\end{tabular}
\caption{Average evaluation of the collected human ratings.}
\label{tab:human-ratings}
\end{table*}

\begin{table*}[t]
\setcode{utf8}
\centering
\begin{tabular}{rrlc}
\hline
\multicolumn{1}{c}{\textbf{Generated Response}}                                                                        & \multicolumn{1}{c}{\textbf{Utterance}}                                                                                                                    &  & \textbf{Emotion} \\ \hline
\small{\<أوه لا! ماذا حدث ؟>}                                                                                          & \small{\<اليوم يوم سيئ جدا>}                                                                                                                              &  & Sadness          \\
\rowcolor[HTML]{EFEFEF} 
\begin{tabular}[c]{@{}r@{}}\small{\<!من الرائع سماع ذلك>}\\ \small{\<ما الذي يجعلك سعيدا جدا حيال ذلك ؟>}\end{tabular} & \small{\<أنا سعيد جدا>}                                                                                                                                   &  & Joy              \\
\small{\<أنا آسف جدا لسماع ذلك. هل أنت بخير ؟>}                                                                        & \small{\<لقد أصبت بفيروس كورونا>}                                                                                                                         &  & Fear             \\
\rowcolor[HTML]{EFEFEF} 
\small{\<إنه لأمر رائع أنك فخور جدا بنفسك>}                                                                            & \small{\<أنا فخور جدا بكوني مهندس>}                                                                                                                       &  & Joy              \\
\small{\<تهانينا! ما هو العمل الشاق ؟>}                                                                                & \small{\<لقد تلقيت ترقية في عملي اليوم>}                                                                                                                  &  & Surprise         \\
\rowcolor[HTML]{EFEFEF} 
\begin{tabular}[c]{@{}r@{}}\small{\<أنا اسف لسماع ذلك>}\\ \small{\<هل كان عيد ميلادك المفضل ؟>}\end{tabular}           & \begin{tabular}[c]{@{}r@{}}\small{\<خلال عيد الميلاد قبل بضع سنوات، لم أحصل>}\\ \small{\<على أي هدايا>}\end{tabular}                                      &  & Sadness          \\
\small{\<أوه ، واو! يجب أن يكون ذلك مزعجا حقا>}                                                                        & \small{\<يستمر جاري في إحداث ضوضاء في الساعة الثانية صباحا>}                                                                                              &  & Anger            \\
\rowcolor[HTML]{EFEFEF} 
\small{\<أوه ، واو ، أنا متأكد من أنك ستكون بخير>}                                                                     & \begin{tabular}[c]{@{}r@{}}\small{\<أختي ستتزوج الأسبوع المقبل . أنا سعيد جدا لها ولكن>}\\ \small{\<في بعض الأحيان أشعر بشيء ثقيل في القلب>}\end{tabular} &  & Sadness          \\ \hline
\end{tabular}
\caption{Examples of responses generated by the BERT2BERT model for multiple utterances with various emotional states and domain contexts.}
\label{tab:examples}
\end{table*}

\subsection{Numerical Evaluation}
Table \ref{tab:metrics} summarizes the perplexity (PPL) and Bilingual Evaluation Understudy (BLEU) scores for the proposed and benchmark models when evaluated on the test set. It is clear from the numerical evaluation results that the proposed BERT2BERT model consistently outperforms the benchmark models. This is reflected through both a lower PPL score and a higher BLEU score.

With EmoPrepend, the addition of supervised information in the form of prepended emotion labels showed performance improvements in comparison to the Baseline model, reflected by an increase in 2.6 BLEU points and a reduction of 14.5 points in the PPL score. Nevertheless, the PPL score of EmoPrepend at 24.1 is still considered high and could potentially lead to sub-optimal performance.  BERT2BERT showed significant performance improvements in comparison to the baseline Seq2Seq Bi-LSTM, highlighted by a much reduced PPL value of 17.0 and an increase in 5 BLEU points. BERT2BERT also achieved better scores than the EmoPrepend model.

The BERT2BERT-UN model resulted in a very high PPL score of 158.9 and very low BLEU score of 0.1. These poor results are due to the nature of transformer networks that require huge amounts of data samples to deliver good performance. The initialization of the BERT2BERT with pre-trained AraBERT weights showed very significant enhancements compared with the uninitialized BERT2BERT-UN model. This performance boost provided by the BERT2BERT model is expected given the fact that AraBERT's initialization parameters have been pre-trained on a massive 24 GB Arabic corpus.

The numerical results achieved by the BERT2BERT model are particularly impressive since, despite the limited size of the  ArabicEmpatheticDialogues dataset, BERT2BERT was able to leverage knowledge transfer through fine-tuning to achieve state-of-art performance on the task of open-domain empathetic response generation in Arabic without requiring additional empathetic samples to train on, or pre-training conversational data.

\subsection{Human Evaluation}
Automated metrics such as PPL and BLEU scores are not sufficient alone to evaluate a model’s ability to exhibit empathetic behavior. Given the unavailability of specific metrics to evaluate empathy in a conversational model, we resort to evaluation based on the judgment of human subjects. Through human evaluation, we can evaluate the emotional communication capability of the models, which is their ability to recognize emotion in the input utterance and generate a suitable expression of emotion in their corresponding response \cite{empathy-evaluation}. To this end, we conducted a survey to collect ratings from 85 native Arabic speakers.

\begin{table*}[ht]
\setcode{utf8}
\centering
\begin{tabular}{rr}
\hline
\multicolumn{1}{c}{\textbf{Generated Response}}   & \multicolumn{1}{c}{\textbf{Utterance}}        \\ \hline
\small{\<أنا سعيد لسماع ذلك>}                     & \small{\<مرحبا كيف حالك ؟>}                   \\
\rowcolor[HTML]{EFEFEF} 
\small{\<ماذا عنك  ؟>}                            & \small{\<ماذا تفعل ؟>}                        \\
\small{\<أوه ، واو. أنا أحب الحيوانات الأليفة>}   & \small{\<ما اسمك ؟>}                          \\
\rowcolor[HTML]{EFEFEF} 
\small{\<لست متأكدا من أين ذهبت. هل ذهبت ؟>}      & \small{\<هل زرت لاس فيجاس من قبل ؟>}          \\
\small{\<هذا يبدو و كأنه عطلة نهاية أسبوع رائعة>} & \small{\<سأذهب للتنزه قليلا و أعود بعد قليل>} \\ \hline
\end{tabular}
\caption{Examples of responses generated by the BERT2BERT model for multiple utterances with neutral emotions.}
\label{tab:examples-neutral}
\end{table*}

\newpage

The raters were shown various utterances and their corresponding responses generated by the Baseline, EmoPrepend, and BERT2BERT models. The BERT2BERT-UN model was excluded from the survey given its poor results in terms of numerical metrics. The raters were asked to evaluate each of the models' ability to show Empathy, Relevance, and Fluency in the generated response. The raters were asked to answer the following questions:

\begin{itemize}
    \item Empathy: Does the generated response show an ability to infer the emotions in the given utterance?
    \item Relevance: How relevant is the generated response to the input utterance?
    \item Fluency: How understandable is the generated response? Is it linguistically correct?
\end{itemize}

For each question, the raters were asked to score the responses of the models on a scale of 0 to 5, where 0 reflects extremely poor performance and 5 reflects excellent performance.

The results of the survey are summarized in Table~\ref{tab:human-ratings}, where we report the average of the collected ratings. The EmoPrepend model showed a higher average score of Empathy and Relevance than the Baseline model. However, these scores are below 3, meaning the EmoPrepend model was seen to deliver below-average performance.

On the other hand, the average ratings of the BERT2BERT model can be considered high and are much superior to both the Baseline and the EmoPrepend models, which indicates BERT2BERT's ability to deliver highly empathetic responses while abiding by linguistic correctness. This is reflected in some examples of the generated responses by BERT2BERT that can be seen in Table~\ref{tab:examples}. The responses demonstrate the model's ability to express empathetic, relevant, and fluent responses when prompted with input utterances with various emotional states and domain contexts, which also proves its ability to handle open-domain conversations.

\subsection{Performance on Inputs with Neutral Emotional States}

Despite the promising results achieved by the BERT2BERT model in generating relevant empathetic responses in open-domain settings, it was shown to poorly handle regular chit-chat utterances with neutral emotions, such as "Hey, how are you?" or "What are you doing?". Instead of providing a regular response, the BERT2BERT model will opt to generate an empathetic response as we show in Table~\ref{tab:examples-neutral}. This issue can be explained by the fact that the model was fine-tuned on a dataset comprised of utterances with pure emotional context and corresponding empathetic responses. Moreso, the AraBERT-initialized parameters did not help mitigate this issue since AraBERT is pre-trained in a self-supervised fashion on news articles and later fine-tuned on a task-specific dataset that does not contain regular chit-chat samples. Thus, it is clear why the BERT2BERT model is not able to handle neutral chit-chat conversations, as it is outside the scope of the training data and the task at hand.

\section{Conclusion}
\label{sec:conc}
In this paper, we address the limitation in resources for Arabic conversational systems, in particular, empathetic conversations. Unlike the English language which has seen great advancements in language generation models due to large corpora and million parameter pre-trained models like GPT, Arabic is considered a low-resource language with limited availability of conversational datasets and pre-trained models for response generation. 

We propose an empathetic BERT2BERT, a transformer-based model, of which the encoder and decoder are warm-started using AraBERT pre-trained parameters and fine-tuned for Arabic empathetic response generation using the ArabicEmpatheticDialogues dataset. By adopting this transfer learning strategy, the proposed BERT2BERT model was able to address the challenges of building an open-domain neural-based empathetic conversational model for a low resource language such as Arabic. BERT2BERT achieved significant performance improvements in comparison to three benchmark models, a baseline Seq2Seq Bi-LSTM model, a Seq2Seq Bi-LSTM model with prepended supervised information about the emotion label during the training process, and a transformer-based encoder-decoder that is not initialized with pre-trained weights. 

The proposed BERT2BERT model achieved a low PPL value of 17.0, a BLEU score of 5.58, and was rated highly by human evaluators with a score of 4.3/5.0, reflecting its ability to generate empathetic, relevant, and fluent responses. Hence, our results show the ability to develop high-performing conversational models in low resource settings by adopting the BERT2BERT strategy.

Despite its high performance in empathetic response generation, BERT2BERT showed a limitation in its ability to handle regular chit-chat conversations with neutral emotional states. To this end, future directions include the development of a strategy that improves the model's ability to determine when an empathetic response is suitable and when it is not.

\section*{Acknowledgments}
This work has been funded by the University Research Board (URB) at the American University of Beirut (AUB).

\bibliography{references,references-relatedwork}

\begin{thebibliography}{31}
\expandafter\ifx\csname natexlab\endcsname\relax\def\natexlab#1{#1}\fi

\bibitem[{Abdelali et~al.(2016)Abdelali, Darwish, Durrani, and
  Mubarak}]{farasa}
Ahmed Abdelali, Kareem Darwish, Nadir Durrani, and Hamdy Mubarak. 2016.
\newblock Farasa: A fast and furious segmenter for arabic.
\newblock In \emph{Proceedings of the 2016 conference of the North American
  chapter of the association for computational linguistics: Demonstrations},
  pages 11--16.

\bibitem[{Ali and Habash(2016)}]{botta}
Dana~Abu Ali and Nizar Habash. 2016.
\newblock Botta: An arabic dialect chatbot.
\newblock In \emph{Proceedings of COLING 2016, the 26th International
  Conference on Computational Linguistics: System Demonstrations}, pages
  208--212.

\bibitem[{Antoun et~al.(2020)Antoun, Baly, and Hajj}]{arabert}
Wissam Antoun, Fady Baly, and Hazem Hajj. 2020.
\newblock {AraBERT}: Transformer-based model for arabic language understanding.
\newblock In \emph{Proceedings of the 4th Workshop on Open-Source Arabic
  Corpora and Processing Tools, with a Shared Task on Offensive Language
  Detection}, pages 9--15.

\bibitem[{ElJundi et~al.(2019)ElJundi, Antoun, El~Droubi, Hajj, El-Hajj, and
  Shaban}]{hulmona}
Obeida ElJundi, Wissam Antoun, Nour El~Droubi, Hazem Hajj, Wassim El-Hajj, and
  Khaled Shaban. 2019.
\newblock hulmona: The universal language model in arabic.
\newblock In \emph{Proceedings of the Fourth Arabic Natural Language Processing
  Workshop}, pages 68--77.

\bibitem[{Fadhil and AbuRa{'}ed(2019)}]{ollobot}
Ahmed Fadhil and Ahmed AbuRa{'}ed. 2019.
\newblock {O}llo{B}ot - towards a text-based {A}rabic health conversational
  agent: Evaluation and results.
\newblock In \emph{Proceedings of the International Conference on Recent
  Advances in Natural Language Processing (RANLP 2019)}, pages 295--303.

\bibitem[{Fan et~al.(2018)Fan, Lewis, and Dauphin}]{topksampling}
Angela Fan, Mike Lewis, and Yann Dauphin. 2018.
\newblock Hierarchical neural story generation.
\newblock In \emph{Proceedings of the 56th Annual Meeting of the Association
  for Computational Linguistics (Volume 1: Long Papers)}, pages 889--898.

\bibitem[{Hijjawi et~al.(2014)Hijjawi, Bandar, Crockett, and Mclean}]{arabchat}
Mohammad Hijjawi, Zuhair Bandar, Keeley Crockett, and David Mclean. 2014.
\newblock {ArabChat:} an arabic conversational agent.
\newblock In \emph{2014 6th International Conference on Computer Science and
  Information Technology (CSIT)}, pages 227--237. IEEE.

\bibitem[{Huang et~al.(2020)Huang, Zhu, and Gao}]{I1}
Minlie Huang, Xiaoyan Zhu, and Jianfeng Gao. 2020.
\newblock Challenges in building intelligent open-domain dialog systems.
\newblock \emph{ACM Transactions on Information Systems (TOIS)}, 38(3):1--32.

\bibitem[{Ippolito et~al.(2019)Ippolito, Kriz, Sedoc, Kustikova, and
  Callison-Burch}]{decodingstrategies}
Daphne Ippolito, Reno Kriz, Joao Sedoc, Maria Kustikova, and Chris
  Callison-Burch. 2019.
\newblock Comparison of diverse decoding methods from conditional language
  models.
\newblock In \emph{Proceedings of the 57th Annual Meeting of the Association
  for Computational Linguistics}, pages 3752--3762.

\bibitem[{Klein et~al.(2017)Klein, Kim, Deng, Senellart, and Rush}]{opennmt}
Guillaume Klein, Yoon Kim, Yuntian Deng, Jean Senellart, and Alexander~M Rush.
  2017.
\newblock Opennmt: Open-source toolkit for neural machine translation.
\newblock In \emph{Proceedings of ACL 2017, System Demonstrations}, pages
  67--72.

\bibitem[{Lewis et~al.(2019)Lewis, Liu, Goyal, Ghazvininejad, Mohamed, Levy,
  Stoyanov, and Zettlemoyer}]{seq2seq-task2}
Mike Lewis, Yinhan Liu, Naman Goyal, Marjan Ghazvininejad, Abdelrahman Mohamed,
  Omer Levy, Ves Stoyanov, and Luke Zettlemoyer. 2019.
\newblock Bart: Denoising sequence-to-sequence pre-training for natural
  language generation, translation, and comprehension.
\newblock \emph{arXiv preprint arXiv:1910.13461}.

\bibitem[{Li et~al.(2017)Li, Su, Shen, Li, Cao, and Niu}]{dailydialog}
Yanran Li, Hui Su, Xiaoyu Shen, Wenjie Li, Ziqiang Cao, and Shuzi Niu. 2017.
\newblock Dailydialog: A manually labelled multi-turn dialogue dataset.
\newblock In \emph{Proceedings of the Eighth International Joint Conference on
  Natural Language Processing (Volume 1: Long Papers)}, pages 986--995.

\bibitem[{Lin et~al.(2020)Lin, Xu, Winata, Siddique, Liu, Shin, and Fung}]{E3}
Zhaojiang Lin, Peng Xu, Genta~Indra Winata, Farhad~Bin Siddique, Zihan Liu,
  Jamin Shin, and Pascale Fung. 2020.
\newblock {CAiRE:} an end-to-end empathetic chatbot.
\newblock In \emph{AAAI}, pages 13622--13623.

\bibitem[{Ma et~al.(2020)Ma, Yang, Du, and Chen}]{I4}
Zhiqiang Ma, Rui Yang, Baoxiang Du, and Yan Chen. 2020.
\newblock A control unit for emotional conversation generation.
\newblock \emph{IEEE Access}, 8:43168--43176.

\bibitem[{Majumder et~al.(2020)Majumder, Hong, Peng, Lu, Ghosal, Gelbukh,
  Mihalcea, and Poria}]{I2}
Navonil Majumder, Pengfei Hong, Shanshan Peng, Jiankun Lu, Deepanway Ghosal,
  Alexander Gelbukh, Rada Mihalcea, and Soujanya Poria. 2020.
\newblock Mime: Mimicking emotions for empathetic response generation.
\newblock In \emph{Proceedings of the 2020 Conference on Empirical Methods in
  Natural Language Processing (EMNLP)}, pages 8968--8979.

\bibitem[{Naous et~al.(2020)Naous, Hokayem, and Hajj}]{naous}
Tarek Naous, Christian Hokayem, and Hazem Hajj. 2020.
\newblock Empathy-driven arabic conversational chatbot.
\newblock In \emph{Proceedings of the Fifth Arabic Natural Language Processing
  Workshop}, pages 58--68.

\bibitem[{Parrott(2001)}]{emotions}
W~Gerrod Parrott. 2001.
\newblock \emph{Emotions in social psychology: Essential readings}.
\newblock Psychology Press.

\bibitem[{Raffel et~al.(2020)Raffel, Shazeer, Roberts, Lee, Narang, Matena,
  Zhou, Li, and Liu}]{seq2seq-task1}
Colin Raffel, Noam Shazeer, Adam Roberts, Katherine Lee, Sharan Narang, Michael
  Matena, Yanqi Zhou, Wei Li, and Peter~J Liu. 2020.
\newblock Exploring the limits of transfer learning with a unified text-to-text
  transformer.
\newblock \emph{Journal of Machine Learning Research}, 21(140):1--67.

\bibitem[{Rashkin et~al.(2019)Rashkin, Smith, Li, and Boureau}]{E2}
Hannah Rashkin, Eric~Michael Smith, Margaret Li, and Y-Lan Boureau. 2019.
\newblock Towards empathetic open-domain conversation models: A new benchmark
  and dataset.
\newblock In \emph{Proceedings of the 57th Annual Meeting of the Association
  for Computational Linguistics}, pages 5370--5381.

\bibitem[{Rothe et~al.(2020)Rothe, Narayan, and Severyn}]{bert2bert}
Sascha Rothe, Shashi Narayan, and Aliaksei Severyn. 2020.
\newblock Leveraging pre-trained checkpoints for sequence generation tasks.
\newblock \emph{Transactions of the Association for Computational Linguistics},
  8:264--280.

\bibitem[{Sharma et~al.(2020)Sharma, Miner, Atkins, and Althoff}]{I3}
Ashish Sharma, Adam Miner, David Atkins, and Tim Althoff. 2020.
\newblock A computational approach to understanding empathy expressed in
  text-based mental health support.
\newblock In \emph{Proceedings of the 2020 Conference on Empirical Methods in
  Natural Language Processing (EMNLP)}, pages 5263--5276.

\bibitem[{Shin et~al.(2020)Shin, Xu, Madotto, and Fung}]{E17}
Jamin Shin, Peng Xu, Andrea Madotto, and Pascale Fung. 2020.
\newblock Generating empathetic responses by looking ahead the user’s
  sentiment.
\newblock In \emph{ICASSP 2020-2020 IEEE International Conference on Acoustics,
  Speech and Signal Processing (ICASSP)}, pages 7989--7993. IEEE.

\bibitem[{Vaswani et~al.(2017)Vaswani, Shazeer, Parmar, Uszkoreit, Jones,
  Gomez, Kaiser, and Polosukhin}]{trans-enc-dec}
Ashish Vaswani, Noam Shazeer, Niki Parmar, Jakob Uszkoreit, Llion Jones,
  Aidan~N Gomez, {\L}ukasz Kaiser, and Illia Polosukhin. 2017.
\newblock Attention is all you need.
\newblock In \emph{Advances in neural information processing systems}, pages
  5998--6008.

\bibitem[{Yal{\c{c}}{\i}n(2019)}]{empathy-evaluation}
{\"O}zge~Nilay Yal{\c{c}}{\i}n. 2019.
\newblock Evaluating empathy in artificial agents.
\newblock In \emph{2019 8th International Conference on Affective Computing and
  Intelligent Interaction (ACII)}, pages 1--7. IEEE.

\bibitem[{Yal{\c{c}}{\i}n(2020)}]{E8}
{\"O}zge~Nilay Yal{\c{c}}{\i}n. 2020.
\newblock Empathy framework for embodied conversational agents.
\newblock \emph{Cognitive Systems Research}, 59:123--132.

\bibitem[{Yal{\c{c}}{\i}n and DiPaola(2018)}]{E15}
{\"O}zge~Nilay Yal{\c{c}}{\i}n and Steve DiPaola. 2018.
\newblock A computational model of empathy for interactive agents.
\newblock \emph{Biologically Inspired Cognitive Architectures}, 26:20--25.

\bibitem[{Yal{\c{c}}{\i}n and DiPaola(2019)}]{mpath}
{\"O}zge~Nilay Yal{\c{c}}{\i}n and Steve DiPaola. 2019.
\newblock M-path: a conversational system for the empathic virtual agent.
\newblock In \emph{Biologically Inspired Cognitive Architectures Meeting},
  pages 597--607. Springer.

\bibitem[{Zhang et~al.(2020{\natexlab{a}})Zhang, Zhao, Saleh, and
  Liu}]{pretraining}
Jingqing Zhang, Yao Zhao, Mohammad Saleh, and Peter Liu. 2020{\natexlab{a}}.
\newblock Pegasus: Pre-training with extracted gap-sentences for abstractive
  summarization.
\newblock In \emph{International Conference on Machine Learning}, pages
  11328--11339. PMLR.

\bibitem[{Zhang et~al.(2018)Zhang, Dinan, Urbanek, Szlam, Kiela, and
  Weston}]{personachat}
Saizheng Zhang, Emily Dinan, Jack Urbanek, Arthur Szlam, Douwe Kiela, and Jason
  Weston. 2018.
\newblock Personalizing dialogue agents: I have a dog, do you have pets too?
\newblock In \emph{Proceedings of the 56th Annual Meeting of the Association
  for Computational Linguistics (Volume 1: Long Papers)}, pages 2204--2213.

\bibitem[{Zhang et~al.(2020{\natexlab{b}})Zhang, Sun, Galley, Chen, Brockett,
  Gao, Gao, Liu, and Dolan}]{dialogpt}
Yizhe Zhang, Siqi Sun, Michel Galley, Yen-Chun Chen, Chris Brockett, Xiang Gao,
  Jianfeng Gao, Jingjing Liu, and William~B Dolan. 2020{\natexlab{b}}.
\newblock {DIALOGPT}: Large-scale generative pre-training for conversational
  response generation.
\newblock In \emph{Proceedings of the 58th Annual Meeting of the Association
  for Computational Linguistics: System Demonstrations}, pages 270--278.

\bibitem[{Zhu et~al.(2015)Zhu, Kiros, Zemel, Salakhutdinov, Urtasun, Torralba,
  and Fidler}]{bookscorpus}
Yukun Zhu, Ryan Kiros, Rich Zemel, Ruslan Salakhutdinov, Raquel Urtasun,
  Antonio Torralba, and Sanja Fidler. 2015.
\newblock Aligning books and movies: Towards story-like visual explanations by
  watching movies and reading books.
\newblock In \emph{Proceedings of the IEEE International Conference on Computer
  Vision}, pages 19--27.

\end{thebibliography}
\bibliographystyle{acl_natbib}



\end{document}